\documentclass[manuscript,screen]{acmart}

\usepackage{listings}
\usepackage{xcolor}

\definecolor{codegreen}{rgb}{0,0.6,0}
\definecolor{codegray}{rgb}{0.5,0.5,0.5}
\definecolor{codepurple}{rgb}{0.58,0,0.82}
\definecolor{backcolour}{rgb}{0.95,0.95,0.92}
\definecolor{linkgray}{rgb}{0.95,0.95,0.95} 

\AtBeginDocument{%
  }

\setcopyright{acmlicensed}
\copyrightyear{2026}
\acmYear{2026}
\acmISBN{978-1-4503-XXXX-X/2018/06}

\begin{document}

\title[AWED-PIPER: Agents, Web apps \& Expert Detectors for FgNER \& PII Protection for 6.6B Speakers]{AWED-PIPER: Agents, Web Applications \& Expert Detectors for Personally Identifiable Information Protection \& Fine-grained Named Entity Recognition  across 36 languages for 6.6 Billion Speakers}


\author{Prachuryya Kaushik}
\email{k.prachuryya@iitg.ac.in}
\orcid{0009-0007-9299-4426}
\affiliation{%
 \institution{Indian Institute of Technology Guwahati}
 \city{Guwahati}
 \state{Assam}
 \country{India}
}

\author{Ashish Anand}
\orcid{0000-0002-0024-3358}
\affiliation{%
 \institution{Indian Institute of Technology Guwahati}
 \city{Guwahati}
 \state{Assam}
 \country{India}
}

\renewcommand{\shortauthors}{Prachuryya Kaushik \& Ashish Anand}


\begin{abstract}
  Named Entity Recognition (NER) and Personally Identifiable Information (PII) anonymization are critical tasks in Natural Language Processing (NLP) for information extraction and privacy preservation. We introduce \textbf{AWED-PIPER}, an open-source framework comprising agentic tools, interactive web applications, and 54 state-of-the-art expert detector models that provide unified Fine-grained Named Entity Recognition (FgNER) and reversible synthetic PII pseudonymization across 36 languages spoken by over 6.6 billion people. The system couples fine-grained multilingual sequence labeling with script-aware regex detectors to identify contextual entities (Person, Location, Organization, Medical) as well as structured technical PII (Emails, native-script Phone Numbers, IP Addresses, Credit Cards). AWED-PIPER offers a dual capability: full FgNER entity extraction and privacy-preserving reversible anonymization with persistent placeholders and de-anonymization dictionary mappings. The suite spans global languages to extremely low-resource vulnerable languages like Bodo, Manipuri, Bishnupriya, and Mizo. The resources can be accessed here: \href{https://github.com/PrachuryyaKaushik/AWED-PIPER}{PII Protector Agentic Tool}, \href{https://github.com/PrachuryyaKaushik/AWED-FiNER}{FgNER Agentic Tool}, \href{https://hf.co/spaces/prachuryyaIITG/AWED_PII_Protector}{PII Web Application}, \href{https://hf.co/spaces/prachuryyaIITG/AWED-FiNER}{FgNER Web Application}, and \href{https://hf.co/collections/prachuryyaIITG/awed-piper}{Edge-deployable Expert Detector Models}.
\end{abstract}





\maketitle

\section{Introduction}
Named Entity Recognition (NER), a fundamental task in Natural Language Processing (NLP), is essential for numerous downstream applications, including information retrieval (IR) \cite{seyler2018information}, web queries \cite{fetahu2021gazetteer}, relation extraction \cite{dagdelen2024structured}, knowledge base construction \cite{christmann2023explainable}, recommendation systems \cite{torbati2021you}, and retrieval-augmented generation \cite{zhang2022end}. Beyond knowledge extraction, identifying entity boundaries plays a critical role in data privacy and Personally Identifiable Information (PII) anonymization. With the rapid deployment of Large Language Model (LLM) agents, sanitizing sensitive personal, medical, organizational, and geographic records prior to external API ingestion has become paramount. However, existing NLP pipelines, IR frameworks, and PII redaction tools predominantly cater to high-resource languages with Latin scripts, thereby marginalizing millions of speakers and leaving low-resource linguistic communities vulnerable to privacy leaks and digital exclusion \cite{xu2023decoupled, zhang2023representation, kaushik2025tafsil}.

\textbf{AWED-PIPER} addresses both granular semantic extraction and cross-lingual privacy preservation by providing a unified, dual-purpose ecosystem across 36 languages. This coverage spans globally spoken languages (such as English, Chinese, Spanish, and Hindi), regional low-resource languages (such as Assamese, Santali, and Odia), and extremely low-resource, vulnerable languages \cite{unesco2017unesco} including Bishnupriya, Bodo, Manipuri, and Mizo. 

\textbf{AWED-PIPER} is a comprehensive suite comprising an \textbf{a}gentic tools, interactive \textbf{w}eb applications, and state-of-the-art \textbf{e}xpert \textbf{d}etector models for \textbf{p}ersonally \textbf{i}dentifiable information \textbf{p}rotection and fine-grained named \textbf{e}ntity \textbf{r}ecognition. The underlying expert detector models are fine-tuned encoders based on the SampurNER \cite{kaushik2026sampurner}, CLASSER \cite{kaushik2025classer}, MultiCoNER2 \cite{fetahu2023multiconer}, FewNERD \cite{ding2021few}, FiNERVINER \cite{kaushik2026finerviner}, FiNE-MiBBiC \cite{kaushik2026fine-mibbic, Kaushik2026fine-mibbic-GitHub}, and APTFiNER \cite{kaushik2026aptfiner} datasets. 

To bridge unstructured entity recognition with privacy engineering, AWED-PIPER couples these expert neural sequence labelers with a script-aware multilingual regex engine. This hybrid design detects contextual entities (Person, Location, Organization, Medical) alongside structured digital PII (Emails, native-script Phone Numbers, IP Addresses, and Credit Cards). Crucially, the framework facilitates \textit{reversible synthetic pseudonymization}, substituting sensitive entities with persistent typed placeholders (e.g., \texttt{[PERSON\_1]}, \texttt{[LOCATION\_1]}) while generating an exact de-anonymization dictionary. These components integrate into callable agentic tools, enabling single-line invocation within autonomous generative AI workflows and LLM agent chains. To the best of our knowledge, this is the first comprehensive framework spanning agentic toolkits, interactive web platforms, and a collection of 54 expert models that simultaneously provides fine-grained entity extraction and multilingual PII pseudonymization for 36 languages serving over 6.6 billion speakers.

\section{The AWED-PIPER Architecture}
AWED-PIPER operates as a unified dual-purpose framework providing both fine-grained named entity recognition and privacy-preserving synthetic pseudonymization across 36 languages.

\subsection{PII Taxonomy Unification}
To support both MultiCoNER2/CLASSER and FewNERD/SampurNER taxonomies, all fine-grained tags are mapped into four primary unstructured PII categories:
\begin{itemize}
    \item \textbf{PERSON}: Maps fine-grained subcategories \textit{Scientist}, \textit{Artist}, \textit{Athlete}, \textit{Politician}, \textit{Actor}, \textit{Scholar}, \textit{Soldier} etc.
    \item \textbf{LOCATION}: Encompasses \textit{HumanSettlement}, \textit{Facility}, \textit{Station}, \textit{GPE}, \textit{Body of Water}, \textit{Transit Infrastructure}, \textit{Buildings} (Airports, Hospitals, Hotels) etc.
    \item \textbf{ORGANIZATION}: Groups \textit{Public/Private Corporations}, \textit{Musical/Sports Groups}, \textit{Educational Institutions}, \textit{Government Agencies} etc.
    \item \textbf{MEDICAL}: Maps clinical mentions including \textit{Diseases}, \textit{Symptoms}, \textit{Medications/Vaccines}, and \textit{Medical Procedures}.
\end{itemize}

\subsection{Hybrid Anonymization Pipeline}
The PII protection module employs a four-phase hybrid architecture:
\begin{enumerate}
    \item \textbf{Expert Model Sequence Labeling}: The multilingual text is processed by language-specific encoder models to identify context-dependent unstructured entities.
    \item \textbf{Multilingual Script-Aware Regex Engine}: A deterministic regex scanner detects structured technical PII (RFC-compliant Emails, IPv4 Addresses, Credit Cards, and Phone Numbers). To support Indian and Asian languages, numeric ranges incorporate native scripts including Devanagari, Bengali, Perso-Arabic, Tamil, Telugu, full-width CJK digits etc.
    \item \textbf{Priority-Based Conflict Resolution \& Span Sanitization}: To prevent NER hallucination on numeric strings, deterministic regex spans take strict precedence over statistical model predictions. Leading and trailing punctuation boundaries (e.g., sentence-ending periods or native virama/purna viram marks) are trimmed from spans to preserve valid syntactic punctuation.
    \item \textbf{Reverse Character Slicing \& Pseudonym Mapping}: Entities are assigned persistent synthetic placeholders (e.g., \texttt{[PERSON\_1]}, \texttt{[LOCATION\_1]}). Replacements are applied in reverse character-offset order (back-to-front) to eliminate UTF-8 byte shift errors across complex multilingual scripts. A reversible JSON de-anonymization dictionary is generated alongside the sanitized text.
\end{enumerate}

\section{Related Works}
Early influential work in NER tools includes the FreeLing suite \cite{carreras2003simple} and German-specific models utilizing Wikipedia-based clusters \cite{chrupala2010named}, both of which established robust baselines for multilingual and language-specific tasks. Widely used NLP tools such as NLTK \cite{loper2002nltk}, CoreNLP \cite{manning2014stanford}, spaCy \cite{choi2015depends}, Flair \cite{akbik2019flair}, and Stanza \cite{qi2020stanza} provide NER pipelines mainly for high-resource languages.  NameTag 1 \cite{strakova2013new} introduces a complex hierarchy for Czech entities, while PolDeepNer \cite{marcinczuk2018recognition} utilizes nested annotations to distinguish granular geographic and personal subtypes. Similarly, the Finnish Tagtools \cite{ruokolainen2020finnish} extend traditional classifications into detailed categories, reflecting a broader trend toward more expressive and precise information extraction models. Although NameTag 3 \cite{strakova-straka-2025-nametag} extended coarse-grained and nested entity recognition to Arabic, Chinese, and 15 European languages, there is still a lack of Fine-grained NER (FgNER) tools spanning various languages, especially for low-resource languages. 

Existing PII anonymization frameworks, such as open-source toolkits (e.g., Microsoft Presidio\footnote{\url{https://presidio.dataprivacystack.org/}}, scrubadub\footnote{\url{https://scrubadub.readthedocs.io/en/stable/}}, ab-ai/pii\_model\footnote{\url{https://huggingface.co/ab-ai/pii_model}}, Roblox\footnote{\url{https://about.roblox.com/newsroom/2025/11/open-sourcing-roblox-pii-classifier-ai-pii-detection-chat}} etc.) and clinical de-identification pipelines, typically combine regular expressions with coarse-grained NER models primarily tailored for English and high-resource Latin scripts. While commercial cloud data protection services offer managed redaction APIs, they remain proprietary, fee-dependent, and lack offline edge-deployment capabilities or support for low-resource languages. AWED-PIPER bridges this gap by unifying 54 fine-grained expert neural detectors with script-aware multilingual regex matching across 36 languages, enabling edge-deployable, reversible synthetic pseudonymization for underserved linguistic communities.

\begin{table}[t!]
\centering
\caption{Performance (Macro-F1) of the best expert models across 22 languages under the MultiCoNER2 \cite{fetahu2023multiconer} taxonomy. To extend language coverage, additional datasets are included: CLASSER \cite{kaushik2025classer}, FiNERVINER \cite{kaushik2026finerviner}, and APTFiNER \cite{kaushik2026aptfiner}.
}
\label{tab:multiconer2_results}
\begin{tabular}{lllccc}
\hline
\textbf{Language} & \textbf{Base Encoder} & \textbf{Dataset} & \textbf{Precision} & \textbf{Recall} & \textbf{Macro-F1} \\
\hline
Assamese   & MuRIL & CLASSER & 74.88 & 75.62 & 75.25 \\
Bengali   & XLM-RoBERTa & MultiCoNER2 & 77.74 & 79.36 & 78.54 \\
Bodo & MuRIL & CLASSER & 73.83 & 76.37 & 75.08 \\ 
Chinese   & XLM-RoBERTa & MultiCoNER2 & 64.66 & 69.42 & 66.95 \\
English   & XLM-RoBERTa & MultiCoNER2 & 78.29 & 80.94 & 79.59 \\
Farsi & XLM-RoBERTa & MultiCoNER2 & 76.05 & 78.86 & 77.43 \\ 
French   & XLM-RoBERTa & MultiCoNER2 & 81.83 & 83.03 & 82.43 \\
German   & XLM-RoBERTa & MultiCoNER2 & 74.51 & 76.29 & 75.38 \\
Hindi & XLM-RoBERTa & MultiCoNER2 & 76.07 & 79.42 & 77.71 \\ 
Italian   & XLM-RoBERTa & MultiCoNER2 & 85.67 & 85.98 & 85.83 \\
Manipuri  & IndicBERTv2 & FiNERVINER & 60.49 & 64.42 & 62.39 \\
Marathi & MuRIL & CLASSER & 79.24 & 81.00 & 80.11 \\ 
Mizo   & XLM-RoBERTa & FiNERVINER & 80.33 & 81.83 & 81.07 \\
Nepali   & MuRIL & CLASSER & 76.92 & 79.50 & 78.19 \\
Portuguese & XLM-RoBERTa & MultiCoNER2 & 80.07 & 81.98 & 81.01 \\ 
Sanskrit   & MuRIL & CLASSER & 77.62 & 78.99 & 78.30 \\
Spanish   & XLM-RoBERTa & MultiCoNER2 & 79.51 & 81.42 & 80.45 \\
Swedish & XLM-RoBERTa & MultiCoNER2 & 85.10 & 84.19 & 84.64 \\ 
Tamil   & XLM-RoBERTa & APTFiNER & 75.51 & 77.28 & 76.38 \\
Telugu  & MuRIL & APTFiNER & 75.72 & 77.26 & 76.49 \\
Ukrainian & XLM-RoBERTa & MultiCoNER2 & 79.78 & 81.51 & 80.61 \\
Urdu & XLM-RoBERTa & CLASSER & 73.48 & 74.92 & 73.93 \\ 
\hline
\end{tabular}
\end{table}

\begin{table}[t!]
\centering
\caption{Performance (Macro-F1) of the best expert models across 27 languages under the FewNERD \cite{ding2021few}. To extend language coverage, additional datasets are included: FiNE-MiBBiC and SampurNER \cite{kaushik2026sampurner}.}
\label{tab:sampurner_results}
\begin{tabular}{lllccc}
\hline
\textbf{Language} & \textbf{Base Encoder} & \textbf{Dataset} & \textbf{Precision} & \textbf{Recall} & \textbf{Macro-F1} \\
\hline
Assamese   & MuRIL & SampurNER & 65.54 & 67.73 & 66.26 \\
Bengali   & MuRIL & SampurNER & 66.54 & 70.08 & 68.26 \\
Bhojpuri & MuRIL & FiNE-MiBBiC & 66.12 & 69.51 & 67.78 \\
Bishnupriya   & MuRIL & FiNE-MiBBiC & 61.85 & 64.13 & 62.97 \\
Bodo   & IndicBERTv2 & SampurNER & 64.17 & 67.14 & 65.62 \\
Chhattisgarhi & MuRIL & FiNE-MiBBiC & 64.40 & 67.87 & 66.09 \\ 
Dogri   & IndicBERTv2 & SampurNER & 60.34 & 64.79 & 62.49 \\
English   & MuRIL & FewNERD & 66.21 & 69.98 & 68.04 \\
Gujarati & IndicBERTv2 & SampurNER & 64.77 & 67.88 & 66.29 \\
Hindi   & IndicBERTv2 & SampurNER & 62.08 & 66.01 & 63.99 \\
Kannada   & IndicBERTv2 & SampurNER & 63.97 & 67.41 & 65.63 \\
Kashmiri & IndicBERTv2 & SampurNER & 60.83 & 64.79 & 62.74 \\
Konkani   & IndicBERTv2 & SampurNER & 61.52 & 65.32 & 63.36 \\
Maithili   & IndicBERTv2 & SampurNER & 60.52 & 64.76 & 62.56 \\
Malayalam & IndicBERTv2 & SampurNER & 62.09 & 65.80 & 63.89 \\
Manipuri   & IndicBERTv2 & SampurNER & 57.68 & 61.14 & 59.35 \\
Marathi   & IndicBERTv2 & SampurNER & 65.09 & 68.47 & 66.73 \\
Mizo & MuRIL & FiNE-MiBBiC & 64.85 & 68.70 & 66.71 \\ 
Nepali   & MuRIL & SampurNER & 67.18 & 70.01 & 68.57 \\
Odia   & IndicBERTv2 & SampurNER & 63.29 & 66.71 & 64.96 \\
Punjabi & IndicBERTv2 & SampurNER & 61.53 & 65.44 & 63.42 \\
Sanskrit   & IndicBERTv2 & SampurNER & 61.48 & 65.19 & 63.27 \\
Santali  & IndicBERTv2 & SampurNER & 51.21 & 52.51 & 52.84 \\
Sindhi & IndicBERTv2 & SampurNER & 60.33 & 62.08 & 61.13 \\
Tamil   & IndicBERTv2 & SampurNER & 62.29 & 65.67 & 63.72 \\
Telugu   & IndicBERTv2 & SampurNER & 62.04 & 65.82 & 63.89 \\
Urdu & IndicBERTv2 & SampurNER & 62.17 & 65.89 & 63.97 \\
\hline
\end{tabular}
\end{table}

\section{Experimental Setup}
The state-of-the-art approach for sequence labeling involves fine-tuning pre-trained encoder models on NER datasets \cite{venkataramana2022hiner, litake2022l3cube, malmasi2022multiconer, mhaske-etal-2023-naamapadam, fetahu2023multiconer, tulajiangbilingual, del2025comparative}. Hence, we fine-tuned IndicBERTv2 (IndicBERTv2-MLM-Sam-TLM) \cite{doddapaneni2023towards}, MuRIL (muril-large-cased) \cite{khanuja2021muril}, and XLM-RoBERTa (XLM-RoBERTa-large) \cite{conneau2020unsupervised} for FgNER using Hugging Face Transformers \cite{wolf2020transformers}. Models were trained for six epochs with batch size 64 using AdamW (learning rate 5e-5, weight decay 0.01) on an NVIDIA A100 GPU. Performance was evaluated using SeqEval metrics, with best models selected by Macro-F1-score \cite{golde2025familarity}.

\section{Results}
For each language, when fine-tuned expert models are available across multiple datasets, the model with the highest Macro-F1 score is preferred for the AWED-FiNER Web Applications and Agentic Toolkits (Table \ref{tab:multiconer2_results} and Table \ref{tab:sampurner_results}). However, all 54 expert models are made available as an open source collection and can be used based on the purpose and the requirement for entity-type granularity.

\section{AWED-PIPER Suite}
The AWED-PIPER suite comprises Agentic Toolkits, Interactive Web Applications, and a collection of 54 compact expert models.

\subsection{Interactive Web Applications} 
Interactive web applications are hosted on Hugging Face Spaces with ZeroGPU dynamically allocated backends:
\begin{enumerate}
    \item \href{https://huggingface.co/spaces/prachuryyaIITG/AWED-FiNER}{\textbf{AWED-FiNER Space}}: Real-time visual FgNER highlighting across 36 languages.
    \item \href{https://huggingface.co/spaces/prachuryyaIITG/AWED_PII_Protector}{\textbf{AWED-PII-Protector Space}}: End-to-end PII masking, synthetic pseudonymization, and JSON dictionary generation.
\end{enumerate}

The \href{https://huggingface.co/spaces/prachuryyaIITG/AWED-FiNER}{AWED-FiNER Hugging Face Space} serves as the backbone for both the interactive web applications and the agentic toolkits. The interactive application, developed using the \texttt{Gradio} framework \cite{abid2019gradio}, enables real-time FgNER visualization across 36 languages. This infrastructure functions as a routing layer that dynamically selects and loads the appropriate compact expert models to minimize resource usage. By providing a unified portal, the space acts as a deployment benchmark for evaluating multilingual performance on complex FgNER taxonomies in resource-constrained environments. \href{https://hf.co/spaces/prachuryyaIITG/AWED_PII_Protector}{AWED-PII-Protector} web application provides reversible synthetic pseudonymization for 36 languages, including four UNESCO-designated \cite{unesco2017unesco} vulnerable languages.

\subsection{Agentic Toolkits} 
Agentic toolkits are implemented for the \texttt{smolagents} and LangChain ecosystems, providing callable tools for LLMs:
\begin{itemize}
    \item \texttt{\href{https://github.com/PrachuryyaKaushik/AWED-FiNER}{AWEDFiNERTool}}: Exposes the \texttt{/predict} API endpoint for entity extraction.
    \item \texttt{\href{https://github.com/PrachuryyaKaushik/AWED-PIPER}{AWEDPIIProtectorTool}}: Exposes the \texttt{/anonymize} API endpoint for privacy filtering before passing prompts to LLMs, with automated de-anonymization restoration helpers.
\end{itemize}

Utilizing this space as callable \texttt{tools}, the \href{https://github.com/PrachuryyaKaushik/AWED-FiNER}{AWED-FiNER agentic toolkit} and \href{https://github.com/PrachuryyaKaushik/AWED-PIPER}{AWED-PII-Protector} implement the \texttt{smolagents} framework to enable automated workflows. Both leverage structured metadata for the 36 supported languages and their fine-grained taxonomies to allow for seamless integration and automatic invocation during inference. This design bridges the gap between static model repositories and autonomous AI applications, ensuring that specialized entity recognition can be effectively integrated into Large Language Models and automated processing pipelines.


\subsection{Multilingual Expert Model Collection}

Our contribution begins with a centralized repository of 54 fine-tuned expert models specialized for FgNER, which incorporates established benchmark datasets: FewNERD \cite{ding2021few}, MultiCoNER2 \cite{fetahu2023multiconer}, CLASSER \cite{kaushik2025classer}, SampurNER \cite{kaushik2026sampurner}, FiNERVINER \cite{kaushik2026finerviner}, APTFiNER \cite{kaushik2026aptfiner}, and FiNE-MiBBiC \cite{kaushik2026fine-mibbic, Kaushik2026fine-mibbic-GitHub}. These \href{https://huggingface.co/prachuryyaIITG/models}{multilingual expert models} target a global audience of 6.6 billion speakers by leveraging multilingual encoder models: IndicBERTv2 \cite{doddapaneni2023towards}, MuRIL \cite{khanuja2021muril}, and XLM-RoBERTa \cite{conneau2020unsupervised}. As each of these compact models has fewer than 355 million parameters, offline deployment is facilitated in resource-constrained scenarios while addressing linguistic disparity across a range of languages, from widely spoken global languages to extremely low-resource and vulnerable languages.



\section{Conclusion}
AWED-FiNER delivers a scalable, efficient FgNER suite for 36 languages and 6.6 billion speakers, with strong support for low-resource and vulnerable languages. AWED-PII-Protector provides efficient and scalable ersonally Identifiable Information
(PII) anonymization for these languages. Its lightweight expert models and agentic tool enable fast, low-memory deployment in resource-constrained environments. To the best of our knowledge, this is the first unified agentic tool, web application, and expert model collection for FgNER task serving 6.6 billion people.

\bibliographystyle{ACM-Reference-Format}
\bibliography{sample-base}

\appendix


\section{Quick Start (Agentic Tools)}

\subsection{Installation}
To install the necessary dependencies, use the following command:
\begin{lstlisting}[language=bash]
pip install smolagents gradio_client
\end{lstlisting}

\subsection{AWED-PII-Protector Tool Usage}
The following snippet demonstrates synthetic pseudonymization and authorized text reconstruction:
\begin{lstlisting}[language=Python]
from tool import AWEDPIIProtectorTool
tool = AWEDPIIProtectorTool(
    space_id="prachuryyaIITG/AWED-PII-Protector"
)
# 1. Anonymize Multilingual Text
raw_text = "Jude Bellingham joined Real Madrid. Email him at jude@realmadrid.es or call +15550199."
result = tool.forward(text=raw_text, language="English")
print("Anonymized Text:\n", result["anonymized_text"])
# Output: "[PERSON_1] joined [ORGANIZATION_1]. Email him at [EMAIL_1] or call [PHONE_1]."

# 2. De-anonymize / Restore Text Downstream
restored_text = tool.deanonymize(result["anonymized_text"], result["deanonymization_map"])
print("\nRestored Text:\n", restored_text)
\end{lstlisting}

\subsection{AWED-FiNER Tool Usage}
The following snippet demonstrates the basic usage of the \texttt{AWEDFiNERTool}:
\begin{lstlisting}[language=Python]
from tool import AWEDFiNERTool
tool = AWEDFiNERTool(
    space_id="prachuryyaIITG/AWED-FiNER"
)
result = tool.forward(
    text="Jude Bellingham joined Real Madrid in 2023.",
    language="English"
)
print(result)
\end{lstlisting}

\subsection{Example usage with Gemini}
The following example illustrates how to integrate the tools with the Google Gemini API:
\begin{lstlisting}[language=Python]
import os
import json
from google import genai
from google.genai import types
from gradio_client import Client
# 1. Setup: Ensure your API key is in the environment
os.environ["GEMINI_API_KEY"] = "YOUR_GEMINI_API_KEY"
client_gemini = genai.Client()
client_gradio = Client("prachuryyaIITG/AWED-FiNER")
# 2. Define the Tool Function
def awed_finer_ner(text: str, language: str):
    """
    Extracts fine-grained named entities from text across 36 languages.
    """
    result = client_gradio.predict(
        text=text,
        language=language,
        api_name="/predict"
    )
    return json.dumps(result)
# 3. Use the Tool with Gemini
response = client_gemini.models.generate_content(
    model="gemini-3-flash-preview",
    contents="Identify the entities in this Mizo sentence: 'Aizawl khaw vawt tak a ni.'",
    config=types.GenerateContentConfig(
        tools=[awed_finer_ner]
    )
)
print(f"Gemini's Response:\n{response.text}")
\end{lstlisting}

\section{Ethical Statement}
We emphasize the preservation of low-resource vulnerable languages (Bodo, Manipuri, Bishnupriya, and Mizo) to prevent digital linguistic extinction. The datasets (SampurNER\footnote{\url{https://huggingface.co/datasets/prachuryyaIITG/SampurNER}}, CLASSER\footnote{\url{https://huggingface.co/datasets/prachuryyaIITG/CLASSER}}, TAFSIL\footnote{\url{https://huggingface.co/datasets/prachuryyaIITG/TAFSIL}}, FiNERVINER\footnote{\url{https://huggingface.co/datasets/prachuryyaIITG/FiNERVINER}}, APTFiNER\footnote{\url{https://huggingface.co/datasets/prachuryyaIITG/APTFiNER}}, FiNE-MiBBiC\footnote{\url{https://github.com/PrachuryyaKaushik/FiNE-MiBBiC}}, MultiCoNERv2\footnote{\url{https://huggingface.co/datasets/MultiCoNER/multiconer_v2}}, FewNERD\footnote{\url{https://huggingface.co/datasets/DFKI-SLT/few-nerd}}) used for fine-tuning the models are released under MIT license\footnote{\url{https://opensource.org/license/MIT}}, CC-BY-4.0\footnote{\url{https://creativecommons.org/licenses/by/4.0/}}, and CC0\footnote{\url{https://creativecommons.org/public-domain/cc0/}} licenses. We did not modify these datasets to correct for potential biases, and we used them as-is. We have cited all the sources of resources, tools, packages, and models used in this work. The AWED-FiNER Web Application\footnote{\url{https://huggingface.co/spaces/prachuryyaIITG/AWED-FiNER}}, Agentic Tool\footnote{\url{https://github.com/PrachuryyaKaushik/AWED-FiNER}} and the collection\footnote{\url{https://huggingface.co/collections/prachuryyaIITG/awed-finer}} of 54 expert models are released MIT license\footnote{\url{https://opensource.org/license/MIT}}.

\end{document}